\documentclass[final,1p,times]{elsarticle}

\usepackage{setspace}
\usepackage{etoolbox}
\AtBeginEnvironment{table}{\singlespacing}
\AtBeginEnvironment{table*}{\singlespacing}
\usepackage{amsmath,amssymb,amsfonts}
\usepackage{graphicx}
\usepackage{booktabs}
\usepackage{multirow}
\usepackage{subcaption}
\usepackage{xcolor}
\usepackage{hyperref}
\usepackage{cleveref}
\usepackage{caption}
\crefname{figure}{Fig.}{Figs.}
\Crefname{figure}{Fig.}{Figs.}
\crefname{table}{Table}{Tables}
\Crefname{table}{Table}{Tables}

\newcommand{\ours}{2D-RoPE-STR}
\newcommand{\ltb}{LTB}
\newcommand{\chunk}{\textsc{chunk-decode-stitch}}

\journal{Pattern Recognition}

\begin{document}

\begin{frontmatter}

\title{Out-of-Length Scene Text Recognition: A Two-Axis Diagnosis and a Training-Free Fix}

\author[inst1]{Zobeir Raisi\corref{cor1}}
\ead{zobeir.raisi@cmu.ac.ir}
\cortext[cor1]{Corresponding author}
\affiliation[inst1]{organization={Electrical Engineering Department, Chabahar Maritime University},
            city={Chabahar},
            country={Iran}}

\author[inst2]{John Zelek}
\affiliation[inst2]{organization={Department of Systems Design Engineering, University of Waterloo},
            city={Waterloo},
            postcode={N2L 3G1},
            state={ON},
            country={Canada}}

\begin{abstract}
Scene Text Recognition (STR) models are trained almost exclusively on word crops of at most 25 characters, yet real deployments (signage, product labels, dense captions) require reading much longer text. This paper diagnoses that failure and then closes it. The diagnosis separates out-of-length failure into two simultaneously extrapolating axes (the encoder's \emph{width} axis and the decoder's \emph{time} axis) and shows that encoder width, not decoder length, is the dominant failure mode. Representation-side fixes bring only partial relief: training-free rotary rescalings recover at most 2--4 points of character error rate (CER), and a weighted fine-tuning recipe recovers 6--8 points while improving standard-benchmark accuracy, yet word accuracy on the Long Text Benchmark (\ltb{}) stays near zero, because the residual gap lies in the decoding mechanism rather than the representation. We then close that gap at inference time, on an unmodified word-level checkpoint: the long image is sliced into overlapping crops at the model's training width, each decoded independently and in-distribution, and the reads stitched by geometry-anchored edit-distance alignment. This procedure reaches 42.79--43.05\% bucket-average word accuracy on \ltb{} across two base checkpoints, matching the published state of the art (41.57\%) and beating it by 11--12 points on the hardest bucket, at wall-clock parity with plain decoding; applied unchanged to the public PARSeq checkpoint it reaches 47.11\%. Once chunking is applied fine-tuning no longer helps: the decoding-side fix alone matches purpose-built architectures. We release the diagnosis harness and implementation.
\end{abstract}


\end{frontmatter}

\section{Introduction}
\label{sec:intro}

Scene Text Recognition (STR) systems are trained almost universally on cropped \emph{words}: the standard synthetic corpora, MJSynth~\citep{jaderberg2016reading} and SynthText~\citep{gupta2016synthetic}, and the standard evaluation suites (IIIT5K~\citep{mishra2012scene}, SVT~\citep{wang2011end}, ICDAR2013~\citep{karatzas2013icdar}, ICDAR2015~\citep{karatzas2015icdar}, SVTP~\citep{quy2013recognizing}, and CUTE80~\citep{risnumawan2014robust}) consist of crops of at most 25--30 characters, and most published architectures inherit a hard decoder length cap and a fixed input resolution matched to that regime. Yet many practical uses of STR (storefront signage, multi-word phrases, dense captions, product labels) require transcribing text considerably longer than a single word. Asked to read a 60- or 100-character phrase, such a model must \emph{extrapolate}: both the encoder, now processing a wider image with more spatial tokens than it saw in training, and the decoder, now emitting more autoregressive steps than its training cap, are pushed out of distribution at once.

This failure is well known empirically: LISTER~\citep{cui2023lister} and the sub-string-matching approach of Du et al.~\cite{du2025smtr} (on the \ltb{} benchmark of 4{,}789 real long-text images we also use) were built to address it. But every existing fix is \emph{decoding}-side: LISTER re-decodes by iteratively matching neighbours, and Du et al.\ decompose recognition into overlapping stitched sub-strings. In NLP, the analogous failure (fixed-context language models failing on longer sequences) is instead addressed on the \emph{representation} side, by rescaling the frequency base of Rotary Position Embedding (RoPE)~\citep{su2024roformer} to keep relative-position geometry well-conditioned beyond training length: position interpolation~\citep{chen2023positional}, NTK-aware scaling, and YaRN~\citep{peng2024yarn}. To our knowledge no prior work asks whether this family transfers to STR, where the setting differs from 1D language modelling in two ways: the encoder is spatial and two-dimensional, and \emph{two} axes (encoder width and decoder time) extrapolate at once and may fail for different reasons.

We build on \ours{}~\citep{raisi2026roperecog}, an encoder-decoder architecture that applies axial 2D-RoPE across the encoder's row and column axes and extends it into decoder cross-attention, while the decoder's own self-attention uses a sinusoidal table; every detail our results depend on is specified in \cref{sec:method-problem,sec:exp-impl}, so the study is self-contained. This gives a minimal-assumption starting point for the length-extrapolation question. The answer carries a lesson beyond STR: RoPE-rescaling transfers only weakly to a 2D visual encoder, because the dominant bottleneck is a \emph{representation} that degrades once the token count leaves its training distribution, not a positional code out of range. Where a length fix should be aimed thus differs in a vision encoder from a language model, and identifying that is as much a contribution as the fix.

Diagnosis is only half the paper. Having localized the residual gap to the decoding mechanism, we close it \emph{entirely at inference time}, with zero training, on an unmodified checkpoint: never present the model an out-of-distribution input at all, but slide overlapping crops at its own training width across the long image, decode each independently and in-distribution, and stitch the reads by geometry-anchored edit-distance alignment. This procedure, \chunk{}, matches and on the hardest bucket exceeds a state-of-the-art long-text architecture trained from scratch, at wall-clock parity with plain decoding. This paper makes four contributions.

\begin{enumerate}
\item We \textbf{decompose} out-of-length failure into an encoder-width axis and a decoder-time axis, evaluating each by holding the other at its training condition. Counter-intuitively, on \ltb{} the width axis alone (an easier sub-task, since only the first 25 characters need be right) already exceeds the decoder axis on the easiest bucket (72.9\% vs.\ 70.6\% CER) and nearly matches it at its worst (84.9\%): encoder-width extrapolation, not decoder length, is the dominant failure mode.
\item We develop \textbf{representation-side fixes and chart their limits}. Adapting three training-free rotary rescalings (NTK-aware, position interpolation (PI), YaRN) to \ours{}'s anisotropic 2D encoder, we find the class only partly effective: NTK, which leaves the local high-frequency dimensions unchanged, is null everywhere (under 1pp CER), while PI and YaRN, which also compress those dimensions, give a consistent 2--4pp CER reduction on the realistic both-axis condition -- localizing the bottleneck as a small recoverable positional-geometry component atop a larger degradation of upstream visual representations. A \textbf{lightweight weighted fine-tuning} recipe (short real-text and long synthetic-text batches, 5{,}000 iterations) goes further, recovering 6--8pp of CER on the hardest \ltb{} buckets and, at an 80/20 short/long weighting, \emph{improving} all six standard STR benchmarks, so it is not a standard-accuracy/length-robustness trade-off.
\item We \textbf{localize the residual gap to the decoding mechanism}. The decoder axis improves under no fine-tuning variant and actively regresses in the 40--75 character range; a stream aimed directly at it (long text squashed to near-illegible size) made every axis worse, a quantified caution against training on synthetic data at the edge of legibility. Two controlled decoder-positional-encoding interventions then both fail (\cref{sec:exp-decoderpe}): rescaling the additive table (PI/YaRN catastrophic, NTK null) and a from-scratch \emph{rotary} decoder that reaches standard-benchmark parity yet leaves the decoder axis unmoved. The two axes thus differ in kind: the width axis is partly representation-recoverable, the decoder axis is not, and the residual gap sits in the decoding \emph{mechanism}, not the positional representation.
\item We introduce \chunk{}, a \textbf{training-free chunked-decoding} algorithm that turns any word-level autoregressive checkpoint into a long-text reader with a single stride hyperparameter and no training. On \ltb{}'s both-axis protocol it reaches 42.79--43.05\% bucket-average word accuracy across two independent base checkpoints, matching the published sub-string-matching state of the art (41.57\%) and exceeding it by 11--12 points on the hardest bucket, at wall-clock parity with plain decoding. We harden this with stride, crop-phase, and base-seed sweeps, a compute-cost analysis, a width-gating deployment rule, and a new synthetic benchmark showing stitching survives long curved text. It is also checkpoint-agnostic: applied unchanged to the public PARSeq checkpoint it reaches 47.11\%, 5.5 points above the state of the art. The two fix families prove \textbf{near-orthogonal}: width-axis fine-tuning does not help, and mildly hurts, once chunking is applied, since chunking feeds the model only in-distribution crops.
\end{enumerate}

All findings are cross-validated on three benchmarks: \ltb{} (real long text, three coarse buckets), a controlled synthetic length curve we built (26--100 characters, real dictionary words, exact length control), and Union14M-B Multi-Words~\citep{jiang2023union14m} as a generalization check. They agree throughout, and the finer synthetic curve additionally reveals the decoder-axis regression \ltb{}'s buckets were too coarse to show.

\section{Related Work}
\label{sec:related}

\subsection{Length Extrapolation in Scene Text Recognition}

The standard STR corpora and the six benchmarks named in \cref{sec:intro} are all built around cropped words of at most 25--30 characters, and Transformer-based recognizers (ViTSTR~\citep{atienza2021vision}, ABINet~\citep{fang2021read}, PARSeq~\citep{bautista2022scene}, and the discriminative--generative model of Yang et al.~\citep{yang2022reading}) inherit both a fixed input resolution and a hard decoder cap from this regime. The Union14M suite~\citep{jiang2023union14m} broadened evaluation along curvature, orientation, artistic fonts, and multi-word structure, but its Multi-Words subset (13.4 average characters) tests multi-word \emph{structure} rather than extreme length, and we use it only as a secondary check.

Two recent surveys track this regime~\citep{afkari2025transformers,kadha2026pixels}, charting the shift from CTC- and rectification-based recognizers (CRNN~\citep{shi2016robust}, ASTER~\citep{shi2018aster}, ESIR~\citep{zhan2019esir}, consolidated by the benchmarking study of Baek et al.~\citep{baek2021what}) to the Transformer designs named above with parallel decoders such as CPPD~\citep{du2025cppd}, and most recently to sub-quadratic models: Mamba-based STR encoders~\citep{zhou2025mambastr,ali2026mamba} target the quadratic-attention cost that also motivates our chunking, but architecturally rather than decoding-side, and report nothing beyond the training-length regime. Dynamic receptive-field adaptation~\citep{tian2024dynamic} is closer to our width-axis diagnosis, varying the effective receptive field with input rather than rescaling position, but is again evaluated only within-distribution. None of this recent architectural work targets out-of-length text directly.

The failure of standard STR models on genuinely long text has been addressed almost exclusively on the \emph{decoding} side. LISTER~\citep{cui2023lister} introduces neighbor decoding, an iterative scheme that predicts characters relative to their neighbours rather than an absolute index. Du et al.~\cite{du2025smtr} propose sub-string matching (SMTR), decomposing a long transcription into overlapping fixed-length sub-strings predicted independently and stitched together, and introduce the \ltb{} benchmark (4{,}789 real images, 26--106 characters) we adopt as our primary set. SVTRv2~\citep{du2025svtrv2} reports a single aggregate long-text number in passing. All of these change how the output is produced or post-processed, not how position is represented internally. To our knowledge no prior STR work studies whether rescaling the \emph{positional encoding itself} (the NLP-established family below) helps, hurts, or is irrelevant here; that is our gap, and the two remain complementary, since a re-decoding scheme could be stacked on any representation-side intervention we study.

A closely related diagnosis comes from outside the Transformer-decoder literature: Diaz et al.~\cite{diaz2021rethinking} report that encoder-decoder line recognizers degrade specifically as image width grows beyond training, and advocate CTC-with-chunking to avoid the problem by construction. This is the same width-axis failure we isolate independently in \cref{sec:exp-phase1}; reached as an architectural workaround rather than a positional-encoding analysis, it complements our finding that the failure is not a positional-frequency artifact (\cref{sec:exp-phase2}). The same brittleness recurs across architectures and modalities: memory-augmented models such as CMAM~\citep{nguyen2019cmam} attack the symptom on long handwritten lines through associative memory, and Garrido-Mu\~{n}oz and Calvo-Zaragoza~\citep{garrido2025generalization} show HTR models generalize poorly to widths far from training. Neither separates the decoder-time axis from the encoder-width axis, the decomposition central to this paper.

\subsection{Length Extrapolation via Positional Encoding in NLP}

In language modelling, models trained with a fixed context window degrade sharply beyond it, and several remedies rescale RoPE~\citep{su2024roformer} to postpone this. Position interpolation~\citep{chen2023positional} linearly compresses test positions back into the training range, trading local resolution for range. NTK-aware scaling, formalized by \citet{peng2024yarn}, instead rescales the frequency base non-uniformly, stretching the highest (most positional) frequencies more so local resolution is preserved while range extends; ALiBi~\citep{press2022train} takes an orthogonal route, replacing rotary position with a linear attention-score penalty. YaRN~\citep{peng2024yarn} combines interpolation and NTK-style rescaling with an attention-temperature correction and reportedly dominates either alone. All three are training-free at their core, which we rely on directly: NTK is the cheapest to adapt and, being closest to a pure frequency-base correction, the most direct test of whether STR's width-axis failure is a positional-frequency phenomenon at all (\cref{sec:exp-phase2}).

This family has developed quickly since, as surveyed by Zhao et al.~\citep{zhao2024lengthsurvey}; Barbero et al.~\citep{barbero2024round} give a mechanistic account of why RoPE's rotation helps attention, and content-conditioned rotary frequencies~\citep{veisi2025carope} are a recent refinement. We evaluate the three original, most widely adopted members (NTK, PI, YaRN) rather than every subsequent variant, since our aim is to test whether the \emph{class} transfers to a 2D visual encoder at all. None of this literature considers a two-dimensional anisotropic encoder coupled to an autoregressive decoder with a \emph{second}, independent extrapolating axis -- the setting that motivates our adaptation and decomposition.

\subsection{2D Rotary Position Embedding and the Base Architecture}

Vision Transformers tokenize an image into a spatial grid of patches~\citep{dosovitskiy2021image}, so a 2D positional code must encode the grid's two axes rather than one sequence index. Axial 2D extensions of RoPE for ViTs~\citep{heo2024rotary,liu2026spiral} rotate half of a token's channels by row index and half by column index, giving an attention score that depends only on relative 2D displacement. Earlier STR-specific encodings addressed the same row/column problem with learnable or content-aware schemes rather than rotary geometry: a 2D positional-embedding Transformer~\citep{raisi2020positional}, the learnable sinusoidal 2LSPE~\citep{raisi2021twolspe}, and a contextual encoding for text in the wild~\citep{raisi2024contextual}. We build on \ours{}~\citep{raisi2026roperecog}, which adapts the axial rotary formulation to STR's anisotropic encoder-decoder setting via an aspect-ratio-matched row/column split and an extension into decoder cross-attention, retaining a sinusoidal table for the decoder's own self-attention. This exposes the two coupled axes we study: the encoder's rotary column axis, extrapolating to more tokens under a wider image, and the decoder's sinusoidal time axis, extrapolating to more steps.

An orthogonal line makes the positional encoding a function of content rather than fixed geometry: SaPE2~\citep{chen2025sape2} conditions position encoding on local semantic context to improve generalization across input scales in vision Transformers. Our rescaling result (\cref{sec:exp-phase2}) finds a purely geometric correction does not fix width-axis extrapolation, because the failure is a representation-distribution problem, not a positional-range one (\cref{sec:discussion}); a content-conditioned encoding in the spirit of SaPE2 is therefore a more promising direction than further geometric rescaling, which we leave to future work.

\section{Method}
\label{sec:method}

The method has two parts, matching the paper's two-step logic. \emph{Part~I} (\cref{sec:method-problem,sec:method-scaling,sec:method-decoder,sec:method-finetune}) formalizes the two-axis problem and develops the representation-side fixes (rotary rescaling, decoder-table extension, fine-tuning) whose limits motivate the rest of the paper. \emph{Part~II} (\cref{sec:method-chunk}) develops \chunk{}, the training-free decoding-side algorithm that closes the residual gap Part~I cannot.

\subsection{Problem Formalization: Two Coupled Extrapolation Axes}
\label{sec:method-problem}

Let $L_{tr}$ be the maximum training transcription length ($L_{tr}=25$, the standard convention). \ours{} processes a fixed $32\times128$ image with a CNN backbone of output stride $s=8$, yielding an $H' \times W' = (H/s)\times(W/s)$ token grid, and the decoder emits up to $L_{tr}$ steps. A test transcription of length $L \gg L_{tr}$ forces \emph{two} simultaneous changes when the image is resized aspect-preservingly to keep characters legible: the encoder receives more column tokens $W'' > W'$, and the decoder must emit more than $L_{tr}$ steps. We call these the \textbf{width axis} and the \textbf{decoder axis}, and their realistic combination \textbf{both axes}.

The two axes extrapolate through different mechanisms in \ours{} and so need not fail, or be fixable, for the same reason. The width axis uses 2D-RoPE: tokens are rotated by their column index at a fixed frequency base, so a token count beyond training means the encoder's self-attention evaluates rotations at frequencies and offsets it never observed. The decoder axis uses an additive sinusoidal table, deterministic and well-defined at any position, so growing it is a lossless buffer extension (\cref{sec:method-scaling}). If the decoder axis nonetheless degrades, the cause cannot be a representation running out of range and must lie in the decoder's learned weights never having produced sequences that long. This asymmetry is why we treat the axes as independently testable hypotheses; \cref{sec:exp} evaluates each in isolation before both together.

We isolate each axis by fixing the other at its training condition: the \textbf{width} condition resizes the image aspect-preservingly (more tokens) but crops the ground truth to the first $L_{tr}$ characters; the \textbf{decoder} condition keeps the standard $32\times128$ image but raises the decode budget; the \textbf{both} condition uses the aspect-preserving image and full transcription, the real end-to-end failure.

\subsection{Width-Axis Scaling: NTK-Aware Anisotropic Rotary Rescaling}
\label{sec:method-scaling}

We adapt NTK-aware scaling~\citep{peng2024yarn} to \ours{}'s anisotropic 2D-RoPE. Because only the column (reading) axis extrapolates under a wider image (the row axis token count $H'$ is fixed by the input height, which does not change for longer text), we rescale only the column-axis frequency base, leaving the row-axis rotation untouched. For an image producing $W''$ column tokens against a training-time count of $W'$, we define the scale factor
\begin{equation}
\gamma(W'') = \max\!\left(\frac{W''}{W'},\, 1\right),
\label{eq:ntk-scale}
\end{equation}
and rescale the column-axis frequency base $b_w$ (nominally $10000$) to $b_w' = b_w \cdot \gamma(W'')$ before computing the rotation angles $\theta^w_i = (b_w')^{-2(i-1)/d}$. This is training-free, computed per image from its token count with no weight change; images no wider than training ($\gamma=1$) recover the original encoding exactly. \Cref{eq:ntk-scale} is the standard NTK-aware construction: it stretches frequencies non-uniformly, so the highest, most position-sensitive frequencies absorb most of the rescaling, in principle preserving local resolution near the training range while extending reach.

For comparison we implement two further column-axis rescalings at the same $\gamma(W'')$. Position interpolation~\citep{chen2023positional} (PI) compresses every frequency uniformly, $\theta^w_i{}' = \theta^w_i / \gamma(W'')$; unlike \cref{eq:ntk-scale} it also rescales the local high-frequency dimensions, trading local resolution for range across the board. YaRN's frequency-domain component~\citep{peng2024yarn} (its ``NTK-by-parts'' correction) blends the two: short-wavelength dimensions keep the original frequency, long-wavelength ones are divided by $\gamma(W'')$ as in PI, with a linear ramp between thresholds ($\beta_{fast}=32$, $\beta_{slow}=1$) in between. We implement this blend only, omitting YaRN's attention-temperature correction, which rescales the attention computation rather than the rotary table. \Cref{sec:exp-phase2} tests all three against the un-rescaled baseline.

\subsection{Decoder-Axis Extension}
\label{sec:method-decoder}

The decoder's positional table is a standard additive sinusoidal buffer~\citep{vaswani2017attention}, deterministic and parameter-free. Extending it to $L_{max} > L_{tr}$ steps is a lossless rebuild: recomputing the table at the larger size leaves every original position numerically unchanged and adds well-defined values at the longer positions. No rescaling analogous to \cref{eq:ntk-scale} applies, since there is no learned frequency and no finite range being exceeded; whatever the decoder axis's failure turns out to be (\cref{sec:exp}), it is therefore a property of the decoder's learned weights and training distribution, not a representation out of range. \Cref{sec:exp-decoderpe} confirms this: neither rescaling this table nor replacing it with a from-scratch rotary decoder moves the decoder axis, whereas both help the encoder's width axis.

\subsection{Fine-Tuning Recipe}
\label{sec:method-finetune}

Since training-free rescaling addresses only the encoding's range and \cref{sec:exp-phase2} finds this only partly sufficient (2--4pp for PI/YaRN, none for NTK, against a much larger failure), we test weight adaptation via short fine-tuning. From a trained checkpoint we extend the decoder table (\cref{sec:method-decoder}) to $L_{max}=100$ and continue training with two weighted-sampled streams, each batch internally uniform-shaped:
\begin{itemize}
\item \textbf{short}: real text from the base checkpoint's own training distribution (Union14M-L~\citep{jiang2023union14m} hard and challenging tiers, $32\times128$), included to guard against forgetting;
\item \textbf{long\_wide}: synthetic text of 26--90 characters rendered at a fixed $512$-pixel canvas width with a dynamically sized font, teaching the model to jointly handle more encoder tokens and longer decode targets in the same example.
\end{itemize}
The \textbf{long\_wide} stream is deliberately minimal: each sample is a uniform-random string over $\{$\texttt{a--z}, \texttt{0--9}$\}$ (no lexicon, so any gain reflects positional/visual generalization rather than language-model memorization), one font (DejaVuSans-Bold) at the largest of six preset sizes fitting the canvas, white text on a random-grey background, no augmentation. This much simpler distribution isolates the shape-level effect of longer, wider text on the positional mechanism without introducing photometric or lexical confounds. (The controlled synthetic \emph{evaluation} curve of Supplementary Sec.~S1 instead uses real dictionary words, since it probes legibility rather than shape.) We sweep the two-stream sampling weight (50/50, 70/30, 80/20 short-weighted) for a fixed 5{,}000-iteration budget, reporting the gain/retention trade-off in \cref{sec:exp-phase3}. We also tried, and report as a negative result, a third stream pairing long synthetic text squashed to a narrow $128$px canvas to target the decoder axis's untrained narrow-image/long-target combination; \cref{sec:discussion} explains why it made every axis worse.

\subsection{Training-Free Chunked Decoding (\chunk{})}
\label{sec:method-chunk}

\begin{figure*}[t]
\centering
\includegraphics[width=0.62\textwidth]{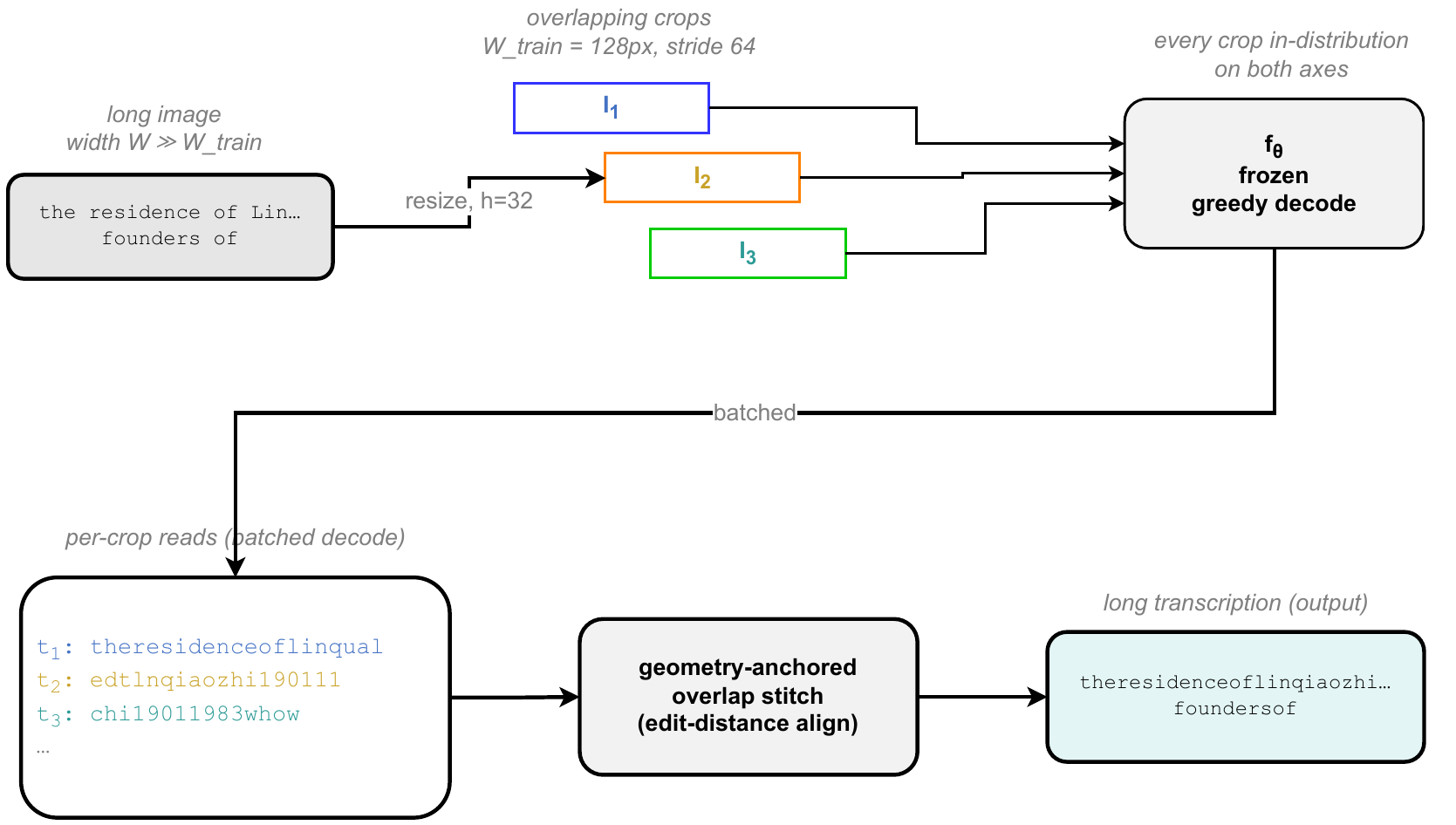}
\caption{\chunk{}: a training-free, inference-time pipeline that turns an
unmodified word-level STR checkpoint $f_\theta$ into a long-text reader. The
wide image is aspect-preserving resized to the training height, sliced into
overlapping crops at exactly the model's training width, each crop decoded
independently by the frozen greedy decoder (so every decode is in-distribution
on both the encoder width axis and the decoder length axis), and the per-crop
reads are reassembled by geometry-anchored edit-distance alignment of their
overlapping substrings. No weights change and no new data is generated; the
only hyperparameter is the crop stride.}
\label{fig:pipeline}
\end{figure*}

Part~I compensates each axis but leaves a residual absolute-accuracy gap on the decoder axis (\cref{sec:method-decoder,sec:exp-decoderpe}). \chunk{} (\cref{fig:pipeline}) takes the opposite route: rather than compensate either axis it \emph{sidesteps both} on an unmodified checkpoint $f_\theta$ with training crop width $W_{\text{train}}$ (128px) and maximum decode length $L_{\text{train}}$ (25). It never presents $f_\theta$ an image wider than $W_{\text{train}}$ nor asks it for more than a short bounded number of decode steps, so every call is in-distribution on both axes.

\subsubsection{Crop Grid}
\label{sec:method-crops}
Given image $I$ of width $W$ (after the same aspect-preserving resize to height 32px used throughout the base pipeline), we place $W_{\text{train}}$-wide crops at left offsets
\[
x_0 = 0,\quad x_i = x_{i-1} + s \ \ (1 \le i < n-1),\quad x_{n-1} = W - W_{\text{train}},
\]
where $s$ is the stride (overlap $= W_{\text{train}} - s$); the first crop is left-aligned and the last right-aligned so the grid covers $I$ regardless of $W \bmod s$, and interior offsets may be shifted by a phase $p \in [0, s)$ without affecting coverage, used only for the robustness check in \cref{sec:exp-hardening}. If $W \leq W_{\text{train}}$, a single crop covers the image and \chunk{} degenerates to plain decoding. Boundary crops narrower than $W_{\text{train}}$ are upsampled to exactly $W_{\text{train}}$, matching training preprocessing. All $n$ crops are batched into one forward pass and decoded in parallel with $f_\theta$'s unmodified greedy decoder, capped at a 30-character per-crop budget (comfortably inside $L_{\text{train}}$).

\subsubsection{Overlap Stitching}
\label{sec:method-stitch}
Adjacent crops $i-1, i$ overlap by $\text{ov}_i = \max(x_{i-1} + W_{\text{train}} - x_i,\ 0)$ pixels, an expected character overlap given the pixels-per-character rate. Because text is not perfectly monospaced and per-crop errors shift boundaries, instead of trusting this estimate exactly, we search a small window of candidate overlap lengths $k$, scoring each by normalized edit-distance similarity between crop $i-1$'s trailing $k$ and crop $i$'s leading $k$ characters, and take the best $k$ above a minimum-similarity threshold; below threshold we fall back to the geometric estimate rather than risk a spurious match. Chunks merge left to right: $\text{acc}_0 = t_0$, $\text{acc}_i = \text{merge}(\text{acc}_{i-1}, t_i, \text{ov}_i)$. This is the overlapping-substring principle of \citet{du2025smtr}, implemented as post-hoc string alignment over an unmodified decoder's outputs rather than a differentiable component of a purpose-built architecture.

\subsubsection{Width-Gating}
\label{sec:method-gate}
\chunk{} should not be applied unconditionally. As \cref{sec:exp-nonhoriz} shows, its resize policy (aspect-preserving, then upsample narrow crops) measurably underperforms the standard training-matched squash resize on images that were never wide to begin with. The deployment rule this motivates is simple: apply plain decoding below a width threshold matched to training width, and \chunk{} only above it -- mirroring the width/decoder-axis split of \cref{sec:method-problem} used throughout the evaluation.

\section{Experiments}
\label{sec:exp}

The head-to-head comparison against every prior method with a published bucketed \ltb{} accuracy (SMTR~\citep{du2025smtr}, LISTER~\citep{cui2023lister}, FocalSVTR, the AR-STR baseline, and PARSeq~\citep{bautista2022scene}) is consolidated in \cref{tab:sota} (\cref{sec:exp-chunk-main}); a reader wanting the bottom line first may read it directly. The intervening subsections build the two-axis diagnosis (\cref{sec:exp-phase1,sec:exp-phase2}) and the representation-side fixes (\cref{sec:exp-phase3,sec:exp-phase4}) whose limits localize the residual gap (\cref{sec:exp-sota}), then the training-free fix that closes it.

\subsection{Datasets}

\textbf{Training.} The base checkpoint is trained on Union14M-L~\citep{jiang2023union14m} (hard and challenging tiers only; see \cref{sec:exp-impl}).

\textbf{Standard benchmarks.} We report the standard six-benchmark STR suite (IIIT5K, SVT, ICDAR 2013, ICDAR 2015, CUTE80, SVTP; all cited in \cref{sec:intro}) to check that a length fix does not silently degrade ordinary word recognition.

\textbf{Long-text evaluation.} Our primary benchmark is \ltb{}~\citep{du2025smtr} (4{,}789 real images, 26--106 characters, all beyond the training cap), bucketed as in prior work into 26--35, 36--55, and 56+ characters (3{,}376 / 1{,}147 / 266 images). Because these buckets are coarse and uneven, we also built a \textbf{controlled synthetic length curve} (real dictionary-word phrases, so failure reflects length not legibility; spaces stripped for exact length control; 11 points from 26 to 100 characters, $n=150$ each, identical decomposition and resize pipeline as \ltb{}). Finally, we use the Union14M-B \textbf{Multi-Words} subset~\citep{jiang2023union14m} as a generalization sanity check; at 13.4 average characters it tests multi-word structure rather than length, and we report it as such.

\subsection{Implementation Details}
\label{sec:exp-impl}

We use the grid \ours{} configuration~\citep{raisi2026roperecog}: a ResNet-50 backbone~\citep{he2016deep} with output stride reduced to 8, a 6-layer encoder with axial 2D-RoPE ($d_{model}=512$, 8 heads, FFN 2048), and a 6-layer decoder using cross-attention 2D-RoPE on the encoder keys and a sinusoidal table over its own sequence. The base checkpoint is trained for 10{,}000 iterations (batch 128, AdamW, 1{,}000-step warmup, mixed precision) on the Union14M-L hard and challenging tiers only -- a reduced-schedule research configuration (80.97\% standard-benchmark average), not a full-convergence production model. This affects absolute numbers but not the relative comparisons the representation-side claims rest on; \cref{sec:exp-phase4} re-examines the absolute numbers on a full-convergence base. Fine-tuning (\cref{sec:method-finetune}) resumes for 5{,}000 iterations (batch 64, lr $10^{-4}$), about 6 minutes on one RTX 4090. All CER figures are corpus-level edit distance normalized by reference length; we report CER rather than exact-match accuracy because accuracy floors to 0.00\% almost everywhere past the training cap (predictions are legible partial reads that drift or loop) and carries no gradient, whereas CER does.

\subsection{Failure Characterization}
\label{sec:exp-phase1}

\Cref{tab:phase1} reports grid \ours{} CER on \ltb{} under the three axis conditions of \cref{sec:method-problem}.

\begin{table}[t]
\centering
\caption{\ours{} (grid) character error rate (\%) on \ltb{}, by extrapolation axis and length bucket. Lower is better.}
\label{tab:phase1}
\begin{tabular}{llr}
\toprule
Axis & Bucket (chars) & CER (\%) \\
\midrule
decoder & 26--35 & 70.63 \\
decoder & 36--55 & 82.73 \\
decoder & 56+    & 84.89 \\
\midrule
width   & 26--35 & 72.90 \\
width   & 36--55 & 73.08 \\
width   & 56+    & 76.15 \\
\midrule
both    & 26--35 & 76.53 \\
both    & 36--55 & 81.57 \\
both    & 56+    & 89.18 \\
\bottomrule
\end{tabular}
\end{table}

The decoder axis shows the expected monotonic cliff, rising roughly 14pp CER from shortest to longest bucket. The width axis is more striking: even on the easiest bucket, where the sub-task is strictly simpler (only the first 25 characters must be right), width-only extrapolation already reaches 72.9\% CER, exceeding the decoder axis's easiest-bucket 70.6\% and approaching its worst-bucket 84.9\%. \textbf{Width-axis extrapolation, not decoder length, is the dominant failure mode}: a grid 2D-RoPE encoder trained only on $128$px ($16$-token) images degrades almost immediately once given a wider image, largely regardless of target length. This attribution is visible only when the axes are evaluated independently, and it redirects where a fix should aim.

\subsubsection{Confirmation on the Controlled Synthetic Length Curve}

The synthetic length curve (\cref{sec:exp}) confirms this cliff with far less bucket-to-bucket noise than \ltb{}'s three uneven buckets: both-axis CER rises monotonically from 64.5\% at 26 characters to 91.0\% at 100 characters, and width-axis CER alone is already 63.8\% at the shortest length point (26 characters), confirming that width dominates decoder length even at finer granularity than \ltb{} offers.

\subsection{Training-Free Rescaling: NTK, PI, and YaRN}
\label{sec:exp-phase2}

Since the failure characterization identifies the width axis as dominant, we test all three training-free rescalings of \cref{sec:method-scaling} (NTK-aware, PI, YaRN) on that axis and the both-axis condition before committing to weight adaptation. \Cref{tab:phase2} reports CER for each.

\begin{table*}[t]
\centering
\caption{Effect of the three training-free column-axis rescalings (\cref{sec:method-scaling}) on \ltb{} CER (\%). $\Delta$ is (baseline $-$ rescaled); positive values indicate rescaling reduced CER.}
\label{tab:phase2}
\begin{tabular}{llrrrr}
\toprule
Axis & Bucket (chars) & Baseline & +NTK ($\Delta$) & +PI ($\Delta$) & +YaRN ($\Delta$) \\
\midrule
width & 26--35 & 72.90 & 73.06 ($-0.16$) & 71.52 ($+1.38$) & 71.55 ($+1.35$) \\
width & 36--55 & 73.08 & 73.56 ($-0.48$) & 74.83 ($-1.75$) & 72.95 ($+0.13$) \\
width & 56+    & 76.15 & 76.90 ($-0.75$) & 77.47 ($-1.32$) & 75.23 ($+0.92$) \\
\midrule
both  & 26--35 & 76.53 & 76.32 ($+0.21$) & 72.68 ($+3.85$) & 73.02 ($+3.51$) \\
both  & 36--55 & 81.57 & 81.16 ($+0.41$) & 78.72 ($+2.85$) & 78.84 ($+2.73$) \\
both  & 56+    & 89.18 & 88.63 ($+0.55$) & 86.99 ($+2.19$) & 87.12 ($+2.06$) \\
\bottomrule
\end{tabular}
\end{table*}

NTK-aware rescaling stays within one point of baseline on every bucket and axis, in inconsistent directions -- indistinguishable from noise: correcting the frequency mismatch while leaving the local high-frequency dimensions untouched does essentially nothing for the width-axis failure. PI and YaRN, which also rescale the local dimensions, differ: on the width-only axis their effect is small and mixed, but on the realistic \textbf{both} condition both give a consistent 2--4pp CER reduction at every bucket, shrinking with length (from $+3.85$/$+3.51$pp at 26--35 to $+2.19$/$+2.06$pp at 56+). The two agree closely at every bucket, supporting a real effect rather than noise: unlike NTK, both trade local resolution for range and recover a similar modest amount by doing so.

This localizes the bottleneck instead of dismissing rescaling: part of the width-axis failure is a recoverable positional-geometry effect (the 2--4pp gain), but the majority is not -- even the best variant recovers under a third of what fine-tuning achieves (\cref{sec:exp-phase3}), and NTK's null shows the more principled resolution-preserving correction fixes nothing. The larger remaining gap is upstream: degraded CNN features or attention diluted over far more tokens than the model was trained to integrate, which no purely positional fix can repair. Weight adaptation (\cref{sec:exp-phase3}), not a larger rescaling grid, is therefore the next step.

\subsection{Fine-Tuning}
\label{sec:exp-phase3}

\Cref{tab:phase3} reports the short/long weighting sweep of \cref{sec:method-finetune} -- standard six-benchmark accuracy and \ltb{} CER.

\begin{table}[t]
\centering
\caption{Fine-tuning weight sweep (short/long). \emph{Top:} standard six-benchmark word accuracy (\%), higher better. \emph{Bottom:} \ltb{} CER (\%), width and both axes, lower better. Baseline is the unfine-tuned checkpoint.}
\label{tab:phase3}
\begin{tabular}{llrrrr}
\toprule
 & & Baseline & 50/50 & 70/30 & 80/20 \\
\midrule
\multicolumn{6}{l}{\emph{Standard six-benchmark word accuracy (\%), higher better}} \\
& IIIT5K  & 87.47 & 85.17 & 87.03 & 88.07 \\
& SVT     & 83.62 & 76.66 & 80.83 & 85.32 \\
& IC13    & 89.36 & 87.19 & 89.06 & 90.34 \\
& IC15    & 71.50 & 62.73 & 67.93 & 72.60 \\
& CUTE80  & 78.82 & 72.92 & 78.47 & 81.25 \\
& SVTP    & 75.04 & 66.82 & 72.09 & 76.74 \\
& \textbf{Avg} & \textbf{80.97} & \textbf{75.25} & \textbf{79.24} & \textbf{82.39} \\
\midrule
\multicolumn{6}{l}{\emph{\ltb{} CER (\%), lower better}} \\
width & 26--35 & 72.90 & 63.04 & 66.51 & 69.79 \\
width & 36--55 & 73.08 & 70.57 & 72.70 & 76.32 \\
width & 56+    & 76.15 & 81.63 & 81.62 & 81.74 \\
both  & 26--35 & 76.53 & 63.42 & 66.96 & 70.34 \\
both  & 36--55 & 81.57 & 70.89 & 73.14 & 76.87 \\
both  & 56+    & 89.18 & 82.02 & 81.91 & 82.18 \\
\bottomrule
\end{tabular}
\end{table}

The sweep traces a single clean trade-off. At \textbf{50/50}, the length gain is largest (roughly 13pp CER on the both axis, easiest bucket) but all six standard benchmarks regress (5.72-point average drop). At \textbf{70/30}, most of the gain is kept (roughly 9 points) while the standard-benchmark cost shrinks by about 70\% (1.73-point drop). At \textbf{80/20}, the checkpoint \textbf{beats baseline on all six standard benchmarks} (+1.42 average -- 80\% real data acting as continued training, the 20\% long stream as a light regularizer) while still cutting both-axis CER by 6--8pp on every bucket. The width-only gain (scored against a 25-character-cropped ground truth, less trustworthy than both) is mostly erased by the 56+ bucket here. We recommend 80/20 as a fix with no standard-benchmark cost; a paper prioritizing the largest length number could report 70/30 with its small quantified cost.

The decoder axis (narrow image, long target, a combination neither training stream covers) does not improve under any variant we tried.

\subsubsection{What Did Not Work: A Third Stream}

We tried a third stream pairing long synthetic text with a narrow $128$px canvas (roughly 1.5 pixels per character) to target the decoder axis's untrained combination directly. At every weight tested (one-third, one-tenth, zero) it made every axis worse or did nothing, including the width and both axes that never touch its image pipeline: text at 1.5 pixels per character is near-unreadable, and predicting an exact label against unreadable input injects gradient noise that degrades training broadly -- a general caution about synthetic data at the edge of legibility. A 4$\times$-longer run (20{,}000 iterations) of the plain two-stream recipe was also worse on eight of nine bucket/axis combinations despite lower training loss, a textbook overfitting signature; we therefore report the 5{,}000-iteration budget as the recipe.

Two secondary confirmations, reported in full in the Supplementary Material (Sec.~S1), reinforce this picture. A finely sampled controlled synthetic length curve (11 points, 26--100 characters) confirms the both-axis fine-tuning gain continuously (12--22 points at every length) and sharpens the decoder-axis behaviour into an \emph{active} regression: fine-tuning is better near the training cap but up to 23 points worse in the 40--75 character range (a repetition-loop signature) before converging back to baseline, exactly the narrow-image/long-target combination that neither fine-tuning stream covers. Separately, on the Union14M-B Multi-Words benchmark the 80/20 fine-tuned checkpoint improves on all seven subsets (33.60$\to$35.47\% average), confirming the recipe behaves as a mild continued-training effect with no observed cost elsewhere.

\subsection{Validation on a Full-Convergence Base}
\label{sec:exp-phase4}

The experiments so far hold the base at the reduced-schedule research model of \cref{sec:exp-impl} (80.97\% standard average), sufficient for every relative comparison but leaving open whether the findings survive on a strong model. We therefore trained a second base from scratch under the \emph{identical} architecture, changing only the data and schedule: all five Union14M-L tiers for 200{,}000 iterations. It reaches \textbf{94.93\% average word accuracy} (IIIT5K 97.7, SVT 97.4, IC13 97.3, IC15 87.9, CUTE80 96.9, SVTP 92.6), comparable to published methods and trained on the same corpus as SMTR, LISTER, and SVTRv2 -- a fair setting to re-ask the question at strength.

\Cref{tab:sota-standard} makes this comparability concrete. The ASTER, SVTR, ABINet, PARSeq, CPPD, LISTER, and MAERec numbers are from Table~3 of \citet{du2025svtrv2}, where all methods are retrained under one common protocol on U14M-Filter (a deduplicated Union14M-L); we report our full-convergence base, trained on the complete corpus, under the same six benchmarks. Its 94.93\% average sits below the strongest recognizers (SVTRv2-B 96.57\%, CPPD/PARSeq 96.40\%) but within 1.4 points of ABINet (95.83\%) and above ASTER (91.70\%) and SVTR (94.58\%): comfortably mid-pack among same-corpus methods, not an outlier propping up the results that follow.

\begin{table}[t]
\centering
\caption{Comparison with published STR methods on the standard six-benchmark suite, all trained on Union14M-L or its filtered variant U14M-Filter~\citep{du2025svtrv2}. Published rows are as reported in Table~3 of \citet{du2025svtrv2}, where every method (SVTRv2 included) is retrained under one common protocol on U14M-Filter; \textbf{Ours} is the full-convergence base of this section, trained on the complete corpus. Best per column in \textbf{bold}.}
\label{tab:sota-standard}
\setlength{\tabcolsep}{3.2pt}
\begin{tabular}{lcccccc|c}
\toprule
Method & IIIT5K & SVT & IC13 & IC15 & CUTE80 & SVTP & Avg \\
\midrule
ASTER~\citep{shi2018aster}         & 96.1 & 93.0 & 94.9 & 86.1 & 92.0 & 87.9 & 91.70 \\
SVTR~\citep{du2022svtr}            & 98.0 & 97.1 & 97.3 & 88.6 & 95.8 & 90.7 & 94.58 \\
ABINet~\citep{fang2021read}        & 98.5 & \textbf{98.1} & 97.7 & 90.1 & 96.5 & 94.1 & 95.83 \\
PARSeq~\citep{bautista2022scene}   & 98.9 & \textbf{98.1} & 98.4 & 90.1 & 98.6 & 94.3 & 96.40 \\
CPPD~\citep{du2025cppd}            & 99.0 & 97.8 & 98.2 & 90.4 & \textbf{99.0} & 94.0 & 96.40 \\
LISTER~\citep{cui2023lister}       & 98.8 & 97.5 & 98.6 & 90.0 & 96.9 & \textbf{94.4} & 96.03 \\
MAERec~\citep{jiang2023union14m}   & \textbf{99.2} & 97.8 & 98.2 & 90.4 & 98.3 & 94.3 & 96.36 \\
SVTRv2-B~\citep{du2025svtrv2}      & \textbf{99.2} & 98.0 & \textbf{98.7} & \textbf{91.1} & \textbf{99.0} & 93.5 & \textbf{96.57} \\
\midrule
Ours (full-convergence base)       & 97.7 & 97.4 & 97.3 & 87.9 & 96.9 & 92.6 & 94.93 \\
\bottomrule
\end{tabular}
\end{table}

\begin{table}[t]
\centering
\caption{\ltb{} CER (\%) on the full-convergence Union14M-L base (94.93\% standard avg), baseline vs.\ the recommended fine-tune (\texttt{v6at\_long}: all-five-tier short stream, 80/20, 15{,}000 iterations). Lower is better.}
\label{tab:phase4-ltb}
\begin{tabular}{llrrr}
\toprule
Axis & Bucket (chars) & Baseline & Fine-tuned & $\Delta$ \\
\midrule
both    & 26--35 & 69.09 & 51.66 & $-17.43$ \\
both    & 36--55 & 79.16 & 64.75 & $-14.41$ \\
both    & 56+    & 89.06 & 81.35 & $-7.71$ \\
\midrule
width   & 26--35 & 65.51 & 51.42 & $-14.09$ \\
width   & 36--55 & 69.68 & 64.52 & $-5.16$ \\
width   & 56+    & 78.87 & 81.05 & $+2.18$ \\
\midrule
decoder & 26--35 & 39.93 & 36.40 & $-3.53$ \\
decoder & 36--55 & 68.01 & 64.08 & $-3.93$ \\
decoder & 56+    & 82.65 & 80.59 & $-2.06$ \\
\bottomrule
\end{tabular}
\end{table}

\begin{figure}[t]
\centering
\includegraphics[width=0.6\columnwidth]{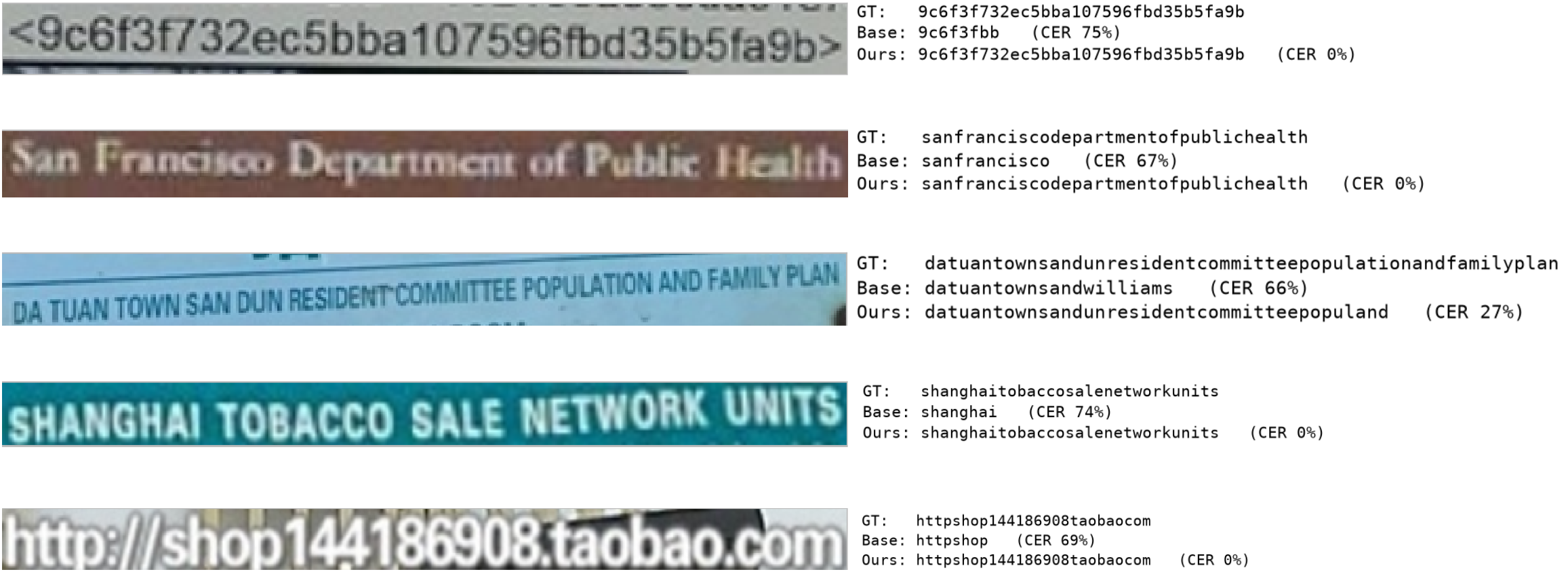}
\caption{Qualitative both-axis reads on \ltb{}: full-convergence base (``Base'') vs.\ the recommended 80/20 fine-tune (``Ours''), using the identical aspect-preserving resize and greedy decode as \cref{tab:phase4-ltb}. On these wide long-text images the baseline consistently stops after the first word or two, the early-termination signature of encoder-width extrapolation (\cref{sec:exp-phase1}); the fine-tune reads through the full line. The last row is a 56+ character example where the fine-tune recovers most but not all of the string (27\% vs.\ 66\% CER), consistent with the residual long-tail gap that \cref{tab:phase4-ltb} and \cref{sec:exp-sota} quantify.}
\label{fig:qual-ft}
\end{figure}

\Cref{tab:phase4-ltb} confirms the width- and both-axis gains are not an artifact of the weak base: on the strong base, fine-tuning cuts both-axis CER by 8 to 17pp, a larger reduction than on the weak base (\cref{tab:phase3}). \Cref{fig:qual-ft} shows the effect on real images: the baseline emits an early end-of-sequence after the first word or two of each wide crop (the truncation signature of width extrapolation), while the fine-tune reads the full line. Three effects nonetheless refine the picture.

The no-cost property of the 80/20 recipe is schedule-dependent. This base is already converged, so it has little under-fit capacity to absorb the 20\% out-of-distribution stream: at 5{,}000 iterations it forgets (standard average 94.93$\to$90.67, Union14M-B 73.88$\to$62.21). Training longer repairs this: at 15{,}000 iterations the perturbation re-integrates and the standard average recovers to 93.83 ($-1.10$), Union14M-B 71.11 ($-2.77$), again nearly free. The weak base behaved oppositely, overfitting the small two-tier set under a longer schedule (\cref{sec:exp-phase3}), so the right fine-tuning length tracks the base's convergence state. The longer strong-base fine-tune also \emph{improves} the decoder axis at every bucket (\cref{tab:phase4-ltb}, e.g.\ 39.93$\to$36.40 at 26--35), which no weak-base variant achieved -- evidence the axis is starved of long-emission signal rather than unfixable, though the gain is small and we do not consider it solved. The accuracy/length trade-off reappears at strength: the 5{,}000-iteration fine-tune reaches the lowest both-axis CER but costs 4 standard-benchmark points, while the 15{,}000-iteration version gives up about 1.6 CER points for near-complete retention (our recommended operating point).

Two further checks support robustness. The training-free rescalings behave as on the weak base (NTK null; PI/YaRN again $-2$ to $-4$pp both-axis CER, YaRN best), so that result is not a base-strength artifact; but applied on top of the fine-tuned checkpoint they hurt (PI collapses word accuracy 10.8\%$\to$0.2\%), since the fine-tuned weights have adapted to un-rescaled positions: fine-tuning subsumes rescaling rather than stacking with it. Repeating the fine-tune under three seeds yields a both-axis spread of about $\pm0.5$ A-Avg (\cref{tab:sota}), well below the effect sizes reported throughout.

\subsection{Localizing the Residual Gap: It Is Decoding-Side}
\label{sec:exp-sota}

The representation-side experiments improve CER substantially, yet absolute exact-match word accuracy on \ltb{} tells a starker story CER hides. On the full-convergence base, fine-tuning lifts word accuracy from essentially zero (A-Avg 0.07) to only 4.30 (recommended) or 5.70 (higher-cost variant) -- below even the weakest purpose-built baseline (AR-STR, 8.51) and far below SMTR (41.57). The full head-to-head against every published \ltb{} method~\citep{du2025smtr} is deferred to \cref{tab:sota} in Part~II, once our own decoding-side fix completes it.

This gap is not base-model strength: our base is competitive on the standard benchmarks (\cref{tab:sota-standard}) and shares the training corpus. It is \emph{decoding-side}. Exact-match accuracy requires the whole long string to be correct, so a model reading half of a 30-character line at $\sim$50\% CER almost never scores a match; and an autoregressive decoder, once past the training length, drifts into repetition loops (AR-STR, the only other purely autoregressive entry in \cref{tab:sota}, is likewise the weakest published method). Every method clearing this bar does so through length-robust \emph{decoding}: SMTR stitches sub-strings, LISTER decodes relative to neighbours, FocalSVTR uses CTC with no step budget to overrun. The residual gap is thus the decoder -- the one component our representation-side fixes leave unchanged, and the target of Part~II.

\subsection{The Decoder Axis Is Not a Positional-Encoding Problem}
\label{sec:exp-decoderpe}

The diagnosis above claims the decoder axis fails in the decoder's learned attention and training distribution, not its positional representation. We test this by attacking the decoder axis from the representation side with the same two tools that help the width axis and showing neither moves it. Numbers are decoder-axis CER on \ltb{} (image squashed to $32\times128$, decode budget 110), full-convergence base.

\emph{Training-free rescaling of the additive table.} Rebuilding the sinusoidal table under PI, NTK, or YaRN, as \cref{eq:ntk-scale} does for the encoder, does not transfer: NTK is neutral-to-harmful, and PI and YaRN are \emph{catastrophic} (\cref{tab:decoderpe}), collapsing into runaway repetition (mean predicted length past 70 characters). An additive absolute code is brittle to re-indexing in a way a relative rotary code is not: compressing its positions hands the decoder a signal it never trained on, with no relative fallback.

\emph{A rotary decoder.} We then replace the additive table entirely with rotary self-attention: queries and keys are rotated by decode step, so decode-time \emph{relative} distance is encoded as on the encoder axes, extrapolates natively, and accepts YaRN. Retraining the base with only this change reaches standard-benchmark parity (94.83\% vs.\ 94.93\%), so the redesign costs nothing where the model already works -- but on the decoder axis it is no better (\cref{tab:decoderpe}), with or without YaRN, before or after fine-tuning. The decisive control is that the rotary base is also slightly worse on the \emph{width} axis, which the decoder's positional encoding cannot influence (the encoder is identical); the small decoder-axis differences are therefore run-to-run variance between two independent 200k trainings, not an effect of the decoder representation.

\begin{table}[t]
\centering
\caption{Decoder-axis CER (\%) on \ltb{}, full buckets ($n=3376/1147/266$, same protocol as \cref{tab:phase4-ltb}), full-convergence base. Neither training-free rescaling of the additive decoder table nor a from-scratch rotary decoder (with or without YaRN, before or after fine-tuning) improves the decoder axis; PI/YaRN on the additive table are catastrophic. The decoder axis is not limited by its positional encoding.}
\label{tab:decoderpe}
\begin{tabular}{lrrr}
\toprule
Decoder positional encoding & L[26,35] & L[36,55] & L$\geq$56 \\
\midrule
Additive table (naive extend) & 39.9 & 68.0 & 82.7 \\
\quad + NTK (training-free)    & 41.9 & 72.2 & 82.7 \\
\quad + PI (training-free)     & 161.1 & 138.8 & 97.3 \\
\quad + YaRN (training-free)   & 202.5 & 180.7 & 115.4 \\
\midrule
Rotary decoder                & 40.7 & 70.4 & 83.5 \\
\quad + YaRN (test-time)       & 46.4 & 70.1 & 81.1 \\
\midrule
Additive table, fine-tuned    & 36.4 & 64.1 & 80.6 \\
Rotary decoder, fine-tuned    & 37.2 & 64.5 & 83.1 \\
\bottomrule
\end{tabular}
\end{table}

Together with \cref{sec:exp-sota}, this is direct, controlled evidence that the axes differ in kind. The width axis is partly a representation problem: encoder-side PI/YaRN (\cref{sec:exp-phase2}) and width-aware fine-tuning (\cref{sec:exp-phase3}) both help it. The decoder axis is not: two independent representation-side interventions leave it unchanged, localizing its failure (and the residual gap of \cref{sec:exp-sota}) to the decoding mechanism. This is what purpose-built decoders target, and what \chunk{} (Part~II) resolves training-free.

\subsection{Chunked Decoding: Setup}
\label{sec:exp-chunk-setup}
Part~II evaluates \chunk{} (\cref{sec:method-chunk}) on the same \ltb{} both-axis protocol and full-convergence base checkpoints as Part~I (\cref{sec:exp-phase4}, 94.9\% standard average), so its numbers are directly comparable to Part~I and to Du et al.~\cite{du2025smtr}. Following the standard protocol under which the published baselines are reported, all transcriptions are scored on the case-insensitive 36-character alphanumeric set (lowercased, spaces and punctuation stripped), so \chunk{}'s stitching operates on an uninterrupted character stream and word segmentation is not part of the task. We use the unmodified base checkpoint as primary, the fine-tuned one for comparison; \chunk{} uses $W_{\text{train}}=128$px, stride 64px (50\% overlap) unless swept, and a 30-character per-crop budget. The layout tests additionally use Union14M-B's Curve and Multi-Oriented subsets~\citep{jiang2023union14m} and a new synthetic long-and-curved generator (\cref{sec:exp-nonhoriz}; Supplementary Sec.~S4).

\subsection{Main Result: Absolute Word Accuracy on \ltb{}}
\label{sec:exp-chunk-main}
\cref{tab:sota} reports the full comparison, completing the deferred table of \cref{sec:exp-sota}. Plain decoding on either checkpoint is non-competitive (A-Avg 0.07--4.30). \chunk{} on the unmodified base reaches A-Avg 42.79, matching Du et al.~\cite{du2025smtr} (41.57) and exceeding it on the hardest bucket ($L\geq56$: 36.84 vs.\ 25.56, $+11.28$). LISTER (26.67) and FocalSVTR (25.59) are beaten decisively on aggregate: they edge ahead on the shortest bucket (51.16/51.04 vs.\ 48.28) but collapse on the two longer ones (2.26\% and 0.38\% at $L\geq56$). PARSeq (0.00) has no long-text mechanism. All of this with \emph{zero} additional training.

\begin{table*}[t]
\centering
\caption{Absolute \ltb{} word accuracy (\%), both-axis, full buckets ($n=3376/1147/266$). The five published rows (SMTR, LISTER, FocalSVTR, AR-STR, PARSeq) are the complete set of prior methods for which a bucketed \ltb{} word accuracy has been reported. Published rows as reported by \citet{du2025smtr}; $^\dagger$\,marks reproductions from \citet{du2025smtr} aligned to a common FocalSVTR encoder, and AR-STR is the autoregressive baseline they report as sinAR. W-Avg is sample-weighted across buckets, A-Avg the unweighted bucket mean. \chunk{} rows use stride 64 (50\% overlap); ``seed2'' is a second, fully independent 200k-iteration base training (\cref{sec:exp-hardening}). $^\ddagger$\,The authors' released \texttt{torch.hub} checkpoint, trained on their own real-data corpus rather than Union14M-L, so this row demonstrates checkpoint-agnosticism of the algorithm rather than a corpus-matched comparison. Best per column in bold.}
\label{tab:sota}
\setlength{\tabcolsep}{4.5pt}
\begin{tabular}{lrrrrr}
\toprule
Method & L[26,35] & L[36,55] & L$\geq$56 & W-Avg & A-Avg \\
\midrule
SMTR~\citep{du2025smtr}        & \textbf{55.48} & 43.68 & 25.56 & \textbf{50.99} & 41.57 \\
LISTER$^\dagger$~\citep{cui2023lister} & 51.16 & 26.59 & 2.26 & 42.56 & 26.67 \\
FocalSVTR                      & 51.04 & 25.37 & 0.38 & 42.08 & 25.59 \\
AR-STR                         & 25.53 & 0.00 & 0.00 & 18.00 & 8.51 \\
PARSeq$^\dagger$~\citep{bautista2022scene} & 0.00 & 0.00 & 0.00 & 0.00 & 0.00 \\
\midrule
Ours, base, plain decode        & 0.21 & 0.00 & 0.00 & 0.15 & 0.07 \\
Ours, +FT, plain decode         & 10.81 & 2.09 & 0.00 & 8.12 & 4.30 \\
\midrule
Ours, base, chunk (s64)      & 48.28 & 43.24 & 36.84 & 46.44 & 42.79 \\
Ours, base, chunk (s96)      & 42.06 & 35.05 & 30.08 & 39.72 & 35.73 \\
Ours, +FT, chunk (s64)       & 47.27 & 41.76 & 36.09 & 45.33 & 41.71 \\
Ours, +FT, chunk (s96)       & 41.02 & 32.35 & 25.19 & 38.06 & 32.85 \\
Ours, base seed2, chunk (s64) & 48.76 & 42.81 & 37.59 & 46.71 & 43.05 \\
\midrule
PARSeq (public)$^\ddagger$, chunk (s64) & 52.67 & \textbf{47.69} & \textbf{40.98} & 50.82 & \textbf{47.11} \\
\bottomrule
\end{tabular}
\end{table*}

\begin{figure*}[t]
\centering
\includegraphics[width=0.74\textwidth]{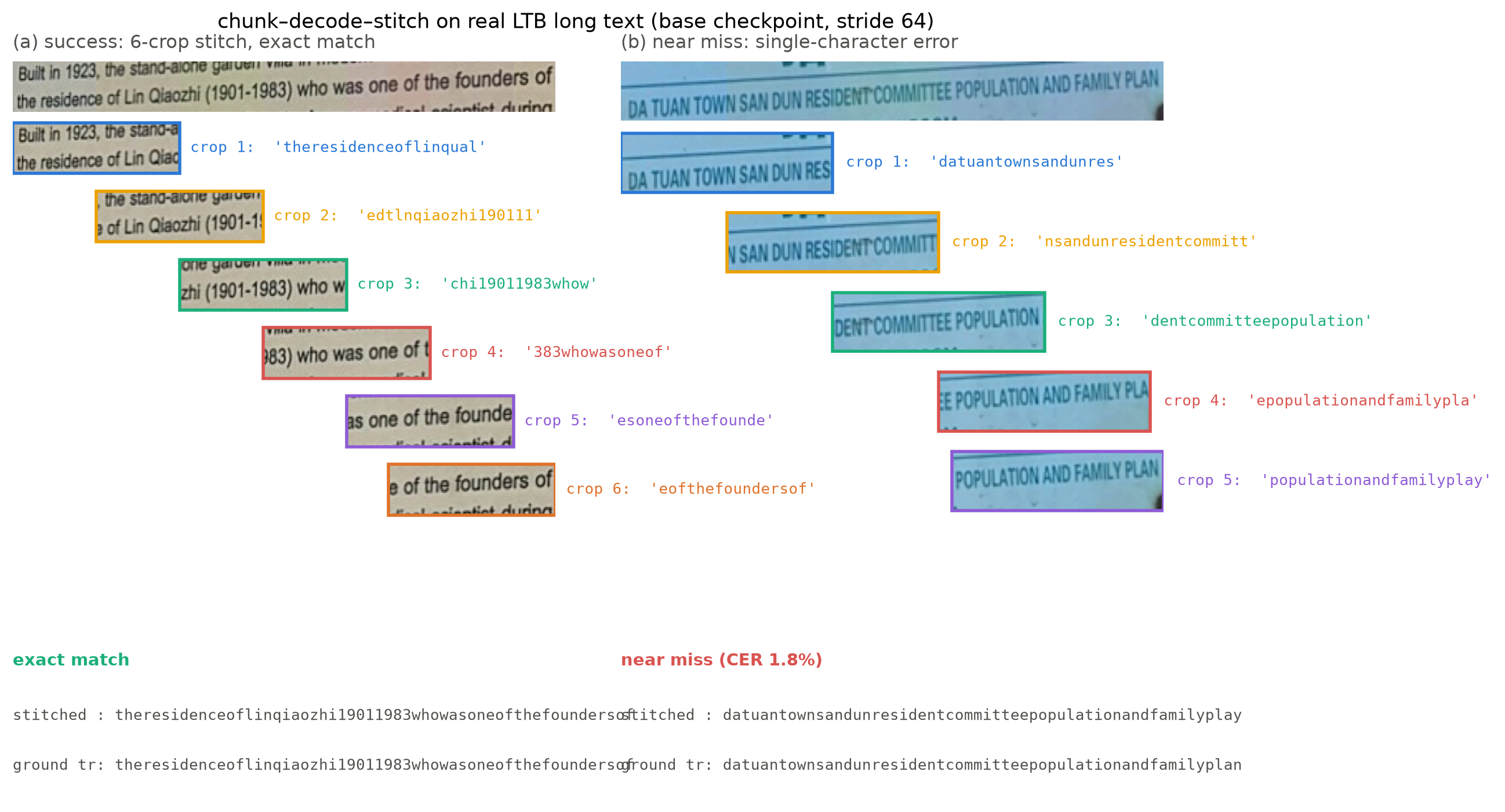}
\caption{\chunk{} on real \ltb{} hardest-bucket ($L\geq56$) examples, base checkpoint, stride 64. Each panel shows the aspect-preserved image strip (top) and the overlapping $128$px crops actually fed to the frozen decoder, staggered downward with their independent in-distribution reads. \textbf{(a)} A six-crop stitch producing an exact match on a multi-line document image, where every crop consistently reads the target line. \textbf{(b)} A representative near miss: a single-character error (\texttt{plan}$\to$\texttt{play}) at the final crop, the interpretable, geometry-explicable error class characteristic of \chunk{}, in contrast to the unstructured drift of plain long-sequence decoding.}
\label{fig:qualitative}
\end{figure*}

Manual inspection of 12 hardest-bucket ($L\geq56$) predictions (\cref{fig:qualitative}) found two exact matches on 60--70 character strings, including a 67-character URL, and confirmed that many remaining errors are not misreads: \ltb{} ground truth contains typos (\texttt{lmage}$\to$\texttt{image}, \texttt{6flfeb}$\to$\texttt{6f1feb}) the model reads \emph{correctly} but which score as failures under exact match. Where it is actually wrong, errors are interpretable and geometry-explicable: a single-character slip at a crop boundary (\cref{fig:qualitative}b), or an over-merge at a repeated substring, not the unstructured drift of plain long decoding.

\textbf{Fine-tuning does not help once chunking is applied.} \cref{tab:sota}'s two \chunk{}-on-fine-tuned rows (41.71/32.85) sit slightly \emph{below} their base-checkpoint counterparts (42.79/35.73) at both strides. \Cref{sec:exp-hardening}'s seed sweep confirms this is not incidental to one fine-tuning run, and \cref{sec:discussion} explains why: chunking feeds the model only in-distribution crops, so the width-axis fine-tuning of Part~I has nothing left to correct.

\textbf{Architecture generality: the algorithm is checkpoint-agnostic.} Everything above uses our own \ours{} checkpoints, so does \chunk{} depend on that architecture? It does not. We applied the identical crop grid and stitcher to the \emph{public} PARSeq checkpoint~\citep{bautista2022scene} from \texttt{torch.hub}, changing only PARSeq's own normalization and tokenizer (its native 25-character cap as the per-crop budget) and touching no weights. Plain decoding scores 0.00 on \ltb{} by construction; chunked, it reaches 52.67/47.69/40.98 (A-Avg 47.11, last row of \cref{tab:sota}) -- 5.5 points above SMTR, ahead on both harder buckets, and above our own chunked checkpoints, consistent with PARSeq being the stronger word-level reader (96.40\% vs.\ our 94.93\%). Its corpus differs from the Union14M-L rows, so we read this as evidence that \chunk{} transfers unchanged to an independent public architecture and amplifies whatever word-level strength the checkpoint has, not as a corpus-controlled comparison.

\subsection{Robustness Hardening}
\label{sec:exp-hardening}
\begin{figure}[t]
\centering
\includegraphics[width=0.6\columnwidth]{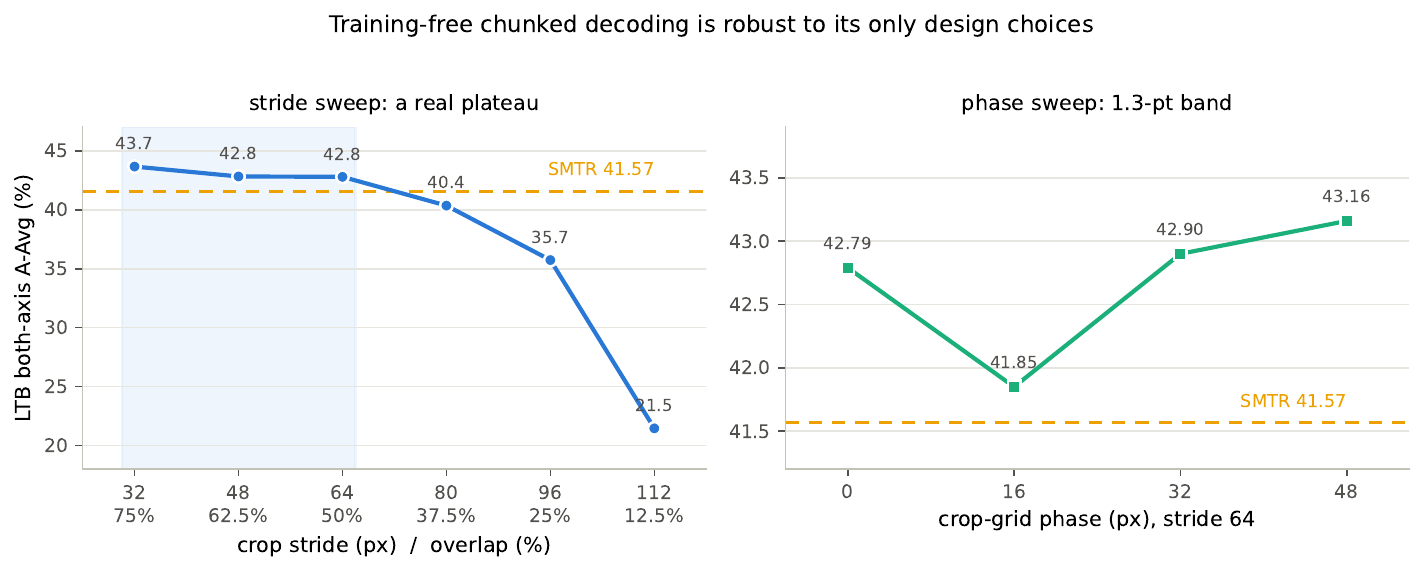}
\caption{Robustness of \chunk{} to its only two design choices, base checkpoint, \ltb{} both-axis A-Avg. \emph{Left:} the crop-stride sweep shows a genuine plateau at $\geq$50\% overlap (shaded), not a cherry-picked optimum, degrading only as overlap falls below it. \emph{Right:} the crop-grid phase sweep spans just 1.3 points. Dashed line: published SMTR A-Avg (41.57).}
\label{fig:robustness}
\end{figure}

A training-free algorithm's headline number is only informative if it is not an artifact of arbitrary choices. \chunk{} is fully deterministic at inference, so instead of averaging over randomness we sweep every design choice and independent training run the result could depend on (\cref{fig:robustness} summarizes the two \chunk{}-internal choices).

\textbf{Stride and crop-grid phase.} A stride sweep from 32 to 112px (75\% to 12.5\% overlap) shows a genuine plateau: strides at or below 64px ($\geq$50\% overlap) sit within 0.9 points of one another, and accuracy degrades smoothly only as overlap falls below that, collapsing at 112px, driven almost entirely by the longest bucket. Shifting the interior crop-grid phase by 0/16/32/48px at the winning 64px stride moves A-Avg by just 1.3 points across four independent phases -- strong evidence, for a method with no weight or sampling randomness, that the result does not depend on where crop boundaries happen to land. Both sweeps are plotted in \cref{fig:robustness}.

\textbf{Fine-tuning seed.} Chunking three independently trained fine-tuning checkpoints (identical recipe, different seed) at stride 64 spreads A-Avg by 9.3 points (far more than \chunk{}'s own design choices), but every seed lands \emph{below} the untouched base's 42.79 (Supplementary Sec.~S2): there is no seed at which this width-axis fine-tuning recipe helps chunked decoding.

\textbf{Base-training seed.} Does the headline 42.79 depend on one lucky base-training run? A second, fully independent 200{,}000-iteration base (identical recipe, different initialization and data-shuffle order) reaches 94.71\% standard average and, chunked at stride 64, A-Avg 43.05 (\cref{tab:sota}, last row) -- a 0.26-point spread from the original, tighter than the phase-shift spread. Both base seeds beat SMTR's 41.57 and win the hardest bucket by 11--12 points: the headline is not an artifact of one training run.

\subsection{Compute Cost}
\label{sec:exp-cost}
\chunk{} decodes more crops per image than plain decoding processes sequences, which could plausibly cost more compute. \cref{tab:cost} reports warm-started wall-clock time per image (5 untimed warmup calls, then CUDA-synchronized timing), $n=150$ per bucket, base checkpoint.

\begin{table}[t]
\centering
\caption{Wall-clock cost per image, base checkpoint, $n{=}150$/bucket.}
\label{tab:cost}
\begin{tabular}{lrrrr}
\toprule
Bucket & Plain (ms) & \chunk{} (ms) & Ratio & Avg.\ crops \\
\midrule
$[26,35]$ & 36.6 & 33.8 & 0.93$\times$ & 6.1 \\
$[36,55]$ & 34.3 & 37.7 & 1.10$\times$ & 7.1 \\
$\geq$56  & 28.3 & 33.4 & 1.18$\times$ & 15.8 \\
\bottomrule
\end{tabular}
\end{table}

\chunk{} is roughly cost-neutral (0.93--1.18$\times$). This is counterintuitive (more crops should cost more), but crops are batched into one parallel forward pass per decode step, while plain decoding runs its longer sequence (up to 110 steps) autoregressively, one step at a time; the two land in the same range. Only the hardest bucket, at nearly 16 crops on average, drifts toward 1.18$\times$. There is no wall-clock objection to deploying \chunk{}.

\subsection{Layout Robustness: Non-Horizontal and Long-Curved Text}
\label{sec:exp-nonhoriz}
Two layout studies, reported in full in the Supplementary Material (Secs.~S3--S4), bound where \chunk{}'s horizontal sliding window applies. First, on Union14M-B's short Curve and Multi-Oriented captions (almost all single-chunk, so this isolates resize policy, not stitching), \chunk{}'s aspect-preserving-then-upsample resize costs 4.0/5.6 points against the training-matched direct squash; images that were never wide to begin with are simply the wrong regime for the long-text pipeline. This is the empirical basis for the width-gating deployment rule of \cref{sec:method-gate}: apply \chunk{} only above a width threshold matched to training width, mirroring the width/decoder-axis split of \cref{sec:method-problem}. Second, because no benchmark combines \emph{long} text with curvature, we build a synthetic one -- real dictionary-word phrases bent along a fixed-wavelength sine midline, so every chunk sees comparable curvature regardless of overall length. Across lengths 30/50/70 and three bend strengths, plain decoding scores 0.00\% at every cell while \chunk{} recovers 84--95\%; curvature costs only a small, length-scaling penalty (about 2 points at length 30, a monotonic 6.7 points at length 70), confirming that stitching survives genuine curvature instead of failing silently.

\section{Discussion}
\label{sec:discussion}

\textbf{Why does positional-frequency rescaling only partly help the width axis?} The width-axis failure appears almost as soon as the encoder sees more tokens than in training, well before long sequences are reached -- the signature of a representation degrading outside its training distribution, not of a positional code running out of resolvable range. Range-limited failure would predict NTK, the variant most directly targeting range, to be the strongest fix; instead it is the weakest, finding nothing to correct because local geometry was never the problem. PI and YaRN help modestly (2--4pp) only because they also compress the local dimensions, making every token look more like a position seen in training. This inverts the NLP context-extension setting that motivated NTK, where representations are not image-derived and NTK is the reported winner.

\textbf{Why does the decoder axis regress under fine-tuning?} The untrained narrow-image/long-target combination sits between our two fine-tuning streams, and filling it directly with a near-illegible narrow-long stream made every axis worse. The lesson is that an image the model cannot possibly read is not a benign extra example -- it degrades gradients for the whole batch, including capabilities the batch does not exercise.

\textbf{Why does a training-free decoding fix succeed where fine-tuning could not?} \chunk{} never asks the decoder to leave its training distribution: every decode is a short in-distribution read, and the long-text problem is pushed entirely into image-space cropping and edit-distance alignment, neither of which has learned parameters. That it closes a gap no representation-side intervention could is confirmation of the diagnosis from the opposite direction -- had the decoder's positional representation been the bottleneck, a method that never touches representation could not have helped.

\textbf{Why does fine-tuning not help, and mildly hurt, once chunking is applied?} \chunk{} feeds the model 128px crops (exactly its training distribution), so no width-axis extrapolation occurs and Part~I's width-axis fix has nothing left to correct, while its shift toward wider training images costs a sliver of narrow-crop quality. The two halves of the paper therefore address different regimes: representation-side extrapolation when the whole image must be decoded at once (irregular layouts, or architectures that cannot crop), chunked decoding whenever cropping is possible, as it is on \ltb{}'s mostly-regular text and plausibly much real signage.

\textbf{Relation to purpose-built decoding architectures.} SMTR~\citep{du2025smtr}, LISTER~\citep{cui2023lister}, and CTC-chunked line recognizers~\citep{diaz2021rethinking} bake the same ``decode short, reassemble long'' insight into a trained architecture. \chunk{} shows the insight alone, with no training, already reaches parity with the strongest of these and surpasses it on a stronger public checkpoint (PARSeq, \cref{tab:sota}) -- not an argument against purpose-built models, which remain preferable when a training budget exists, but evidence that \chunk{} is a strong free baseline any new method should beat and a deployable fix when a checkpoint cannot be retrained.

\section{Conclusion}
\label{sec:conclusion}

This work set out to understand why word-trained Scene Text Recognition models fail on longer text, and it arrives at a conclusion with both a diagnostic and a practical component. Diagnostically, out-of-length failure is not one problem but two coupled ones, and the two are unequal: the encoder's \emph{width} axis, not the decoder's time axis, is the dominant and previously unattributed bottleneck, and it behaves like a representation leaving its training distribution rather than a positional code running out of range. Practically, the part of the gap that survives representation-side repair is decoding-side, and it can be closed with no training at all by keeping every decode in-distribution and moving the long-text problem into image-space cropping and string alignment.

The main strength of this result is leverage: the decoding-side fix is checkpoint-agnostic and training-free, so a practitioner holding any existing word-level model (including public checkpoints such as PARSeq that score zero on long text by construction) can reach parity with purpose-built long-text architectures without retraining, new data, or an architecture change, at wall-clock parity with plain decoding. The field benefits three ways: researchers gain a strong, essentially free baseline any new long-text method should beat before claiming a training-based contribution; a sharper target, since the diagnosis relocates the width-axis bottleneck to the visual representation and so redirects effort from positional-encoding tweaks toward encoders that stay well-conditioned as the token count grows; and two dead ends reported honestly as negative results: NTK-aware rescaling is a clean null here, and injecting near-illegible narrow-and-long synthetic data degrades training instead of filling the decoder-axis gap.

The work also has clear weaknesses. The decoder axis remains the least well addressed by representation-side means: it regresses under fine-tuning on the reduced-schedule base, improves only modestly on the strong base, and resisted a direct attempt to target it. The stitcher has a known over-merging failure at repeated substrings that we document qualitatively but do not quantify as its own error category. Our robustness evidence covers curvature and resize policy but not dense multi-line or vertical layouts, a plausibly harder regime for a horizontal window. The 36-character protocol strips spaces, case, and punctuation, so recovering word boundaries in space-delimited text lies outside the task as evaluated. Finally, our headline comparisons place our own harness against \emph{published} numbers (the standard convention, but worth restating), and one comparison mixes a full-Union14M-L base against methods retrained on the filtered variant, so the corpora are close but not byte-identical.

These weaknesses map onto future work. The clearest next step is a curriculum-style fine-tuning schedule introducing the narrow-and-long decoder condition only after the two working streams converge, paired with synthetic data of controlled rather than unreadable per-character legibility. On the decoding side, a confidence-weighted or lightly learned stitcher could remove the repeated-substring over-merge, and extending the sliding window from a horizontal strip to a two-dimensional crop grid would bring multi-line and vertical layouts, with an explicit word-boundary step, into scope. And because the width-axis diagnosis concerns two-dimensional visual encoders rather than STR specifically, testing whether the same representation-versus-position distinction governs length extrapolation in adjacent dense-prediction settings (document understanding, full-page OCR, dense captioning) would establish how far the lesson generalizes.

\singlespacing
\section*{Code and data availability}
All code, including the diagnosis harness, the three rescalings (NTK, PI, YaRN), the fine-tuning recipe, \chunk{} (with the public-PARSeq adapter), the synthetic generators, and every figure/table script, is available at \url{https://github.com/zobeirraisi/chunk-decode-stitch}. All datasets are public (\cref{sec:exp}).

\section*{CRediT authorship contribution statement}
\textbf{Zobeir Raisi:} Conceptualization, Methodology, Software, Validation, Formal analysis, Writing -- original draft, Writing -- review \& editing. \textbf{John Zelek:} Writing -- review \& editing.

\section*{Declaration of competing interest}
The authors declare no known competing financial interests or personal relationships that could have appeared to influence the work reported in this paper.

\section*{Declaration of generative AI and AI-assisted technologies in the writing process}
During the preparation of this work, the authors used generative AI and AI-assisted tools to assist with language editing, translation, grammar correction, formatting, and readability improvement. The authors reviewed and edited the content as needed and take full responsibility for the content of the submitted manuscript. No generative AI tools were used to generate scientific results, data, tables, figures, or conclusions.

\bibliographystyle{elsarticle-num-names}
\bibliography{references}

\clearpage
\setcounter{section}{0}
\setcounter{figure}{0}
\setcounter{table}{0}
\renewcommand{\thesection}{S\arabic{section}}
\renewcommand{\thefigure}{S\arabic{figure}}
\renewcommand{\thetable}{S\arabic{table}}

\section*{Supplementary Material}

This supplement collects secondary confirmations and robustness studies that support, but are not required by, the argument of the main paper. Section and table numbers of the form 4.x, Table~x, etc.\ refer to the main paper; supplementary items are numbered S1, S2, \dots. Every experiment here uses the same checkpoints, benchmark, and evaluation protocol described in the main paper.

\section{Deferred Confirmations: Synthetic Length Curve and Multi-Words}
\label{sec:s-deferred}

The main paper's failure characterization and fine-tuning results are established on the three coarse length buckets of the Long Text Benchmark (\ltb{}). This section confirms both findings continuously, on a finely sampled controlled synthetic length curve, and checks that the fine-tuned checkpoint does not lose ground on a short-text multi-word benchmark.

\begin{figure}[t]
\centering
\includegraphics[width=0.72\textwidth]{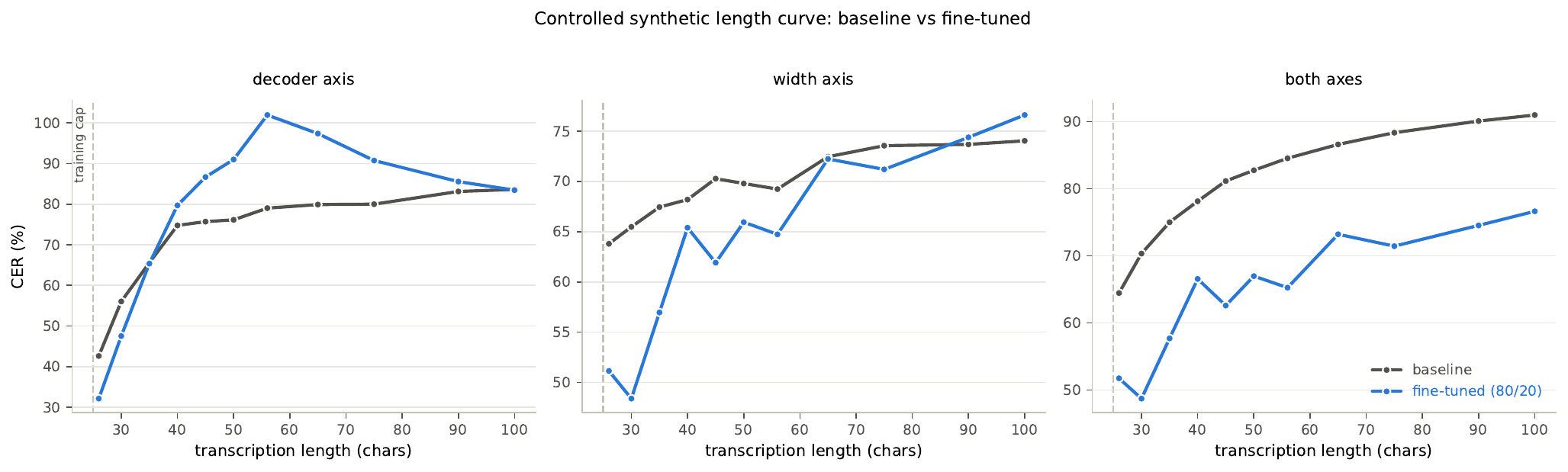}
\caption{Controlled synthetic length curve (26--100 characters, 11 points, $n{=}150$/point): baseline vs.\ 80/20 fine-tuned CER (\%), decoder / width / both axes. The fine-tuning gain on the both axis is sustained across the full range; the width-axis gain crosses back above baseline near 90--100 characters; the decoder axis regresses in the 40--75 character range.}
\label{fig:s-synth}
\end{figure}

\Cref{fig:s-synth} plots the controlled synthetic length curve for both the baseline and the recommended 80/20 fine-tuned checkpoint, across all three axes, at all 11 length points from 26 to 100 characters.

This finer-grained curve confirms both the diagnosis's and the fine-tuning's headline findings continuously rather than at three coarse points, and surfaces one finding \ltb{} was too coarse to show clearly: on the \textbf{both} axis, fine-tuning cuts CER by 12--22 percentage points at every one of the 11 length points, a sustained gain across the full 26--100 character range. On the \textbf{width} axis alone, the fine-tuned checkpoint is better through roughly 75 characters, then crosses back above the baseline by 90--100 characters, the same ``gain erased at the long tail'' pattern the 56+ \ltb{} bucket had already hinted at, now visible as an actual crossover point. On the \textbf{decoder} axis, however, fine-tuning is not merely neutral: it is better near the training cap (26--35 characters, 8--10 percentage points better than baseline), then \emph{actively worse} from roughly 40--75 characters (up to 23 percentage points worse at length 56, where fine-tuned CER briefly exceeds 100\%, i.e.\ more inserted or substituted characters than the reference is long, a repetition-loop signature), converging back to roughly tied with baseline by 90--100 characters. This is a real limitation of the recipe: a plausible cause is that neither fine-tuning stream ever pairs a narrow image with a long target, so the decoder's behaviour in that combination is left unconstrained by the fine-tuning, consistent also with why the targeted third fine-tuning stream reported in the main paper failed to fix it directly.

\Cref{tab:s-multiword} reports the Union14M-B Multi-Words secondary benchmark, run unmodified on both checkpoints as a generalization sanity check (not a length-extrapolation result, given its short average label length).

\begin{table}[t]
\centering
\caption{Union14M-B word accuracy (\%), all seven subsets, baseline vs.\ 80/20 fine-tuned. A generalization sanity check, not a length benchmark (average label length 13.4 characters).}
\label{tab:s-multiword}
\begin{tabular}{lrrr}
\toprule
Subset & Baseline & Fine-tuned & $\Delta$ \\
\midrule
Artistic        & 35.33 & 36.11 & $+0.78$ \\
Contextless     & 41.72 & 43.90 & $+2.18$ \\
Curve           & 27.25 & 28.98 & $+1.73$ \\
General         & 59.23 & 60.34 & $+1.11$ \\
Multi-Oriented  & 14.61 & 17.97 & $+3.36$ \\
Multi-Words     & 33.41 & 37.27 & $+3.86$ \\
Salient         & 23.66 & 23.72 & $+0.06$ \\
\midrule
\textbf{Avg}    & \textbf{33.60} & \textbf{35.47} & \textbf{+1.87} \\
\bottomrule
\end{tabular}
\end{table}

The fine-tuned checkpoint improves on all seven subsets, Multi-Words included, on a benchmark neither fine-tuning stream nor the standard six-benchmark suite was built to touch, consistent with the fine-tuning's reading that the 80/20 weighting behaves as a mild continued-training effect with no observed cost elsewhere.

\section{Fine-Tuning-Seed Sweep for Chunked Decoding}
\label{sec:s-robustness}

The main paper summarizes \chunk{}'s robustness to its two design choices (crop stride and crop-grid phase) with a figure, and reports the base-training-seed replicate. This section reports the remaining, fine-tuning-seed sweep (base checkpoint, \ltb{} both-axis A-Avg).

\textbf{Fine-tuning seed.} \Cref{tab:s-ftseed} chunks three independently trained fine-tuning checkpoints (identical recipe, different training seed) at stride 64. The spread (9.3 points) is far larger than the stride or phase spreads (fine-tuning seed matters in a way \chunk{}'s own design choices do not), but every seed lands \emph{below} the untouched base's 42.79: there is no seed at which this width-axis fine-tuning recipe helps chunked decoding.

\begin{table}[t]
\centering
\caption{Fine-tuning-seed sweep, stride 64, \ltb{} both-axis A-Avg. All three underperform the untouched base (42.79).}
\label{tab:s-ftseed}
\begin{tabular}{lrrr}
\toprule
Seed & 0 & 1 & 2 \\
\midrule
A-Avg & 34.01 & 30.01 & 39.29 \\
\bottomrule
\end{tabular}
\end{table}

\section{Non-Horizontal Layout and the Width-Gating Rule}
\label{sec:s-nonhoriz}

\ltb{} is overwhelmingly regular horizontal text (signage, URLs, phrases); a sliding horizontal window is a strong prior specifically for this layout. We test Union14M-B's Curve and Multi-Oriented subsets (full subsets, $n=2426$/$1369$) to probe whether the algorithm generalizes to curved and rotated \emph{captions}. Because these are short-caption crops, 99.4\%/98.8\% of samples resolve to a single chunk -- this is not yet a test of multi-chunk stitching under curvature (\cref{sec:s-curved} tests that directly) but does isolate \chunk{}'s own resize policy (aspect-preserving, then upsample if narrower than $W_{\text{train}}$) against the standard training-matched direct squash.

\begin{table}[t]
\centering
\caption{Non-horizontal-layout resize-policy comparison, Union14M-B, base checkpoint. Almost all samples are single-chunk; the gap isolates resize policy, not stitching.}
\label{tab:s-nonhoriz}
\begin{tabular}{lrrrr}
\toprule
Subset & Plain acc & \chunk{} acc & Plain CER & \chunk{} CER \\
\midrule
Curve           & 76.46\% & 72.42\% & 9.55\%  & 12.91\% \\
Multi-Oriented  & 67.13\% & 61.50\% & 18.56\% & 26.11\% \\
\bottomrule
\end{tabular}
\end{table}

\chunk{}'s resize policy measurably costs accuracy on these subsets (4.0/5.6-point drop) despite the near-total absence of multi-chunk stitching, most plausibly because two sequential resizes (aspect-preserve, then upsample) lose more fine detail than a single direct squash, and that detail loss matters more when text is already small, curved, or rotated within its crop. This resolves into an actionable deployment rule rather than a vague caveat, and is the empirical basis for the width-gating rule stated in the main paper's method section: gate \chunk{} on image width, applying it only above a threshold matched to training width, mirroring the width/decoder-axis split. Curve and Multi-Oriented images are short captions that were never wide to begin with; applying the long-text pipeline to them is simply the wrong tool, not evidence against the method for the regime it targets.

\section{Long, Curved Text: Does Stitching Survive Curvature?}
\label{sec:s-curved}

The open question (whether chunk \emph{stitching} degrades on long text that bends across chunk boundaries, as opposed to \ltb{}'s overwhelmingly straight signage) could not be answered by \cref{sec:s-nonhoriz}, since Union14M-B's curved captions are essentially never wide enough to require more than one chunk. No existing benchmark combines long text with curvature, so we build one: real dictionary-word phrases rendered at their natural content-fit width (using the same controlled-length generator as the synthetic length-curve experiment of \cref{sec:s-deferred}), then bent along a sine midline with a \emph{fixed pixel wavelength} ($\approx$160px, approximately one chunk width) rather than a single B\'ezier arc stretched over the whole variable-length canvas. This design choice matters: an initial attempt reusing a whole-image B\'ezier midline produced curvature that was visually imperceptible within any single 128px chunk once stretched over a 650--1200px long-text image, which would have silently failed to test the question at all. A fixed pixel wavelength guarantees every chunk sees comparable curvature regardless of overall image length.

\Cref{tab:s-curved} reports word accuracy and CER at three lengths (30/50/70 characters, spanning \ltb{}'s bucket range) and three bend strengths (straight/mild/strong), $n=150$ samples per cell, exact synthetic ground truth, base checkpoint, stride 64.

\begin{table}[t]
\centering
\caption{Long+curved synthetic stitching test, base checkpoint, stride 64, $n{=}150$/cell.}
\label{tab:s-curved}
\begin{tabular}{llrrrr}
\toprule
Len. & Bend & Plain acc & Plain CER & \chunk{} acc & \chunk{} CER \\
\midrule
30 & straight & 0.00\% & 68.08\% & 94.67\% & 0.42\% \\
30 & mild     & 0.00\% & 68.95\% & 92.67\% & 0.53\% \\
30 & strong   & 0.00\% & 67.08\% & 92.67\% & 0.51\% \\
50 & straight & 0.00\% & 84.55\% & 87.33\% & 0.88\% \\
50 & mild     & 0.00\% & 84.93\% & 90.00\% & 0.76\% \\
50 & strong   & 0.00\% & 85.25\% & 86.00\% & 0.83\% \\
70 & straight & 0.00\% & 90.39\% & 90.67\% & 0.39\% \\
70 & mild     & 0.00\% & 90.44\% & 86.00\% & 0.60\% \\
70 & strong   & 0.00\% & 90.36\% & 84.00\% & 0.50\% \\
\bottomrule
\end{tabular}
\end{table}

Plain decoding fails completely at every condition (0.00\% word accuracy), matching the real-\ltb{} base-checkpoint pattern -- length alone is already fatal, and curvature is not even the deciding factor there. \chunk{} recovers 84--95\% accuracy at every condition, including the strongest curvature tested. Curvature does cost something, and the cost scales with length, consistent with more chunks producing more stitching boundaries at which local curve-induced misalignment can compound: at length 30, straight-to-strong costs roughly 2 points (94.67$\to$92.67, plausibly within $n{=}150$ sampling noise); at length 70 it is a real, monotonic 6.7-point cost (90.67$\to$84.00). Even at the single worst observed condition, \chunk{}'s 84.00\% remains far ahead of plain decoding's 0.00\%. This closes the caveat: chunk stitching survives long, genuinely curved text, with a small, quantified, length-scaling accuracy cost rather than the silent catastrophic failure \cref{sec:s-nonhoriz} alone could not rule out.
\end{document}